\documentclass[10pt,twocolumn,letterpaper]{article}

\usepackage{cvpr}
\usepackage{times}
\usepackage{epsfig}
\usepackage{graphicx}
\usepackage{amsmath}
\usepackage{amssymb}
\usepackage{bm}

\usepackage[pagebackref=true,breaklinks=true,letterpaper=true,colorlinks,bookmarks=false]{hyperref}

\cvprfinalcopy % *** Uncomment this line for the final submission

\def\cvprPaperID{32} % *** Enter the CVPR Paper ID here

\usepackage[linesnumbered, ruled]{algorithm2e}
\usepackage{blindtext}
\usepackage[small, tight, center]{subfigure}

\newcommand{\name}{WiCapture\xspace}

\newenvironment{Itemize}%
{\vspace{-2mm}
\begin{itemize}%[leftmargin=12pt]
\setlength{\itemsep}{0pt}%
\setlength{\topsep}{0pt}%
\setlength{\partopsep}{0pt}%
\setlength{\parskip}{0pt}}%
{\end{itemize}
\vspace{-2mm}}

\newenvironment{Glo}%
{\vspace{-2mm}
\begin{itemize}%[leftmargin=12pt]
\setlength{\itemsep}{0pt}%
\setlength{\topsep}{0pt}%
\setlength{\partopsep}{0pt}%
\setlength{\parskip}{0pt}}%
{\end{itemize}
\vspace{0mm}}

\newenvironment{Enumerate}%
{\vspace{-2.3 mm}
\begin{enumerate}%[leftmargin=12pt]
\setlength{\itemsep}{0pt}%
\setlength{\topsep}{0pt}%
\setlength{\partopsep}{0pt}%
\setlength{\parskip}{0pt}}%
{\end{enumerate}
\vspace{-2.3 mm}} 

\usepackage{verbatim}

\newcommand{\substart}{\vspace{-4mm}}
\newcommand{\subend}{\vspace{-2mm}}
\ifcvprfinal\fi
\begin{document}

%%%%% MANI commands
% to reduce space above and below equations
\setlength{\abovedisplayskip}{4pt}
\setlength{\belowdisplayskip}{4pt}

%%%% to reduce space for sections
%\newcommand{\msection}{\vspace{-2mm} \section}
%\newcommand{\msubsection}{\vspace{-3mm} \subsection}
%\newcommand{\msubsubsection}{\vspace{-2mm} \subsubsection}

%%%%% END MANI commands

%%%%%%%%% TITLE
\title{Position Tracking for Virtual Reality Using Commodity WiFi}

\author{Manikanta Kotaru, Sachin Katti \\
Stanford University\\
{\tt\small \{mkotaru, skatti\}@stanford.edu }
}
%\author{First Author\\
%Institution1\\
%Institution1 address\\
%{\tt\small firstauthor@i1.org}
%% For a paper whose authors are all at the same institution,
%% omit the following lines up until the closing ``}''.
%% Additional authors and addresses can be added with ``\and'',
%% just like the second author.
%% To save space, use either the email address or home page, not both
%\and
%Second Author\\
%Institution2\\
%First line of institution2 address\\
%{\tt\small secondauthor@i2.org}
%}

%%%Improvements from Sigcomm wesite
%\clubpenalty=10000 
%\widowpenalty=10000 

\maketitle
\thispagestyle{empty}

%%%%%%%%% ABSTRACT
\begin{abstract}
Today, experiencing virtual reality (VR) is a cumbersome experience which either requires dedicated infrastructure like infrared cameras to track the headset and hand-motion controllers (\eg Oculus Rift, HTC Vive), or provides only 3-DoF (Degrees of Freedom) tracking which severely limits the user experience (\eg Samsung Gear VR). To truly enable VR everywhere, we need position tracking to be available as a ubiquitous service. This paper presents WiCapture, a novel approach which leverages commodity WiFi infrastructure, which is ubiquitous today, for tracking purposes. We prototype WiCapture using off-the-shelf WiFi radios and show that it achieves an accuracy of $0.88$ cm compared to sophisticated infrared-based tracking systems like the Oculus, while providing much higher range, resistance to occlusion, ubiquity and ease of deployment.

\end{abstract}

%%%%%%%%% BODY TEXT
\section{Introduction}

Immersive experiences like virtual reality (VR) require accurate tracking of the headset and other accessories like hand-motion controllers. Current commercial tracking systems like Oculus Rift~\cite{rift} and HTC Vive~\cite{vive} are outside-in where the tracking is performed using infrastructure external to the VR accessories. The external infrastructure is specialized and typically uses infrared (IR) cameras along with sensors on the headset to perform the tracking. These systems are very accurate but have the following limitations:
\vspace{-3mm}
\begin{Itemize}
  \item{They require installing specialized hardware and dedicated infrastructure wherever user wants to experience VR. So if a user wishes to use VR headsets anywhere in her home, one would need IR cameras everywhere.}
  \item{These systems are not occlusion resistant. For example, if the camera is blocked by furniture or if the user turns away from the camera, then the tracking fails.}
  \item{These systems have limited range, typically around $2$ m in front of the camera~\cite{dk2Video}.}
  \end{Itemize}

A competing technology to provide position tracking is inside-out position tracking found in systems like the Microsoft Hololens~\cite{holo}. These systems use cameras (both RGB and depth sensing) and implement vision based tracking algorithms on the headset. These systems are both accurate and infrastructure-free, however they come with certain limitations. Specifically, they significantly increase the complexity of the headset since they need to have several cameras as well as complex algorithms running on the headset to provide tracking. Further they are not robust, tracking fails in environments with transparent or texture-less objects (\eg a white wall)~\cite{patent:20160131761}. Finally and most importantly, these systems cannot be used for tracking peripherals such as hand-motion controllers; the complexity of inside-out tracking is too high to be implemented on such peripherals which are meant to be lightweight and cheap.

\begin{figure}
\begin{center}
\includegraphics[width=0.8\linewidth]{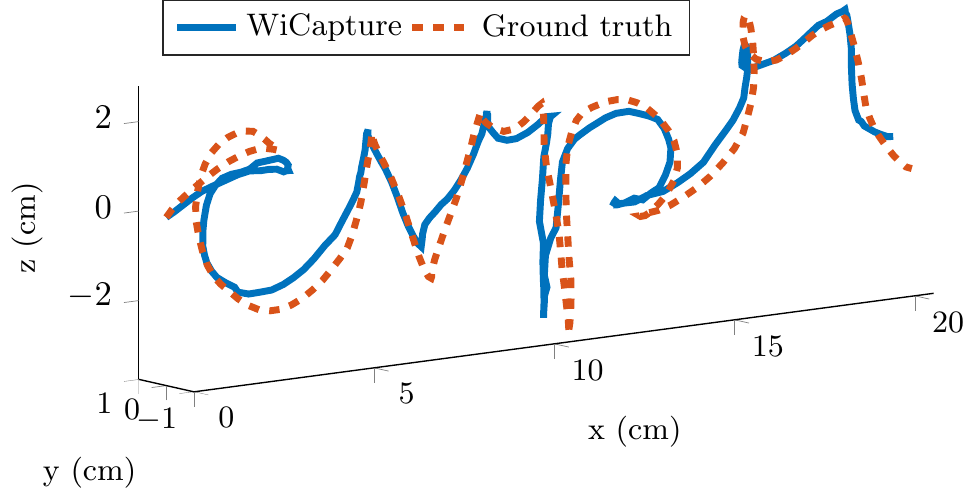}
\end{center}
\caption{Solid blue path estimated by \name is compared against the dotted red ground truth trajectory.}
\label{fig:trace}
\end{figure}

In this paper, we present WiCapture, a novel VR position tracking system which addresses the above limitations of existing systems. WiCapture is a WiFi based position tracking system. Headsets transmit standard WiFi packets which are received by standard WiFi access points (APs). The WiFi APs receive metadata from each WiFi packet reception called Channel State Information (CSI) which encodes the transformation the environment has induced upon the transmitted WiFi signals. WiCapture invents novel algorithms that mine the CSI metadata to recover the position of the headset accurately. It has the following properties:
\begin{Itemize}
  \item{WiCapture does not require special hardware, it uses commodity APs that can be bought in retail stores.}
  \item{WiCapture is occlusion resistant, it continues to work even when the APs and headsets are occluded due to furniture or other objects in between.}
  \item{WiCapture has larger range and operates across rooms.}%, it works as long as there is a WiFi AP in the room.}
  \item{It is insensitive to room illumination or texture and it can work in the dark. Further, headset complexity is minimal, all the headset needs is a standard WiFi chip.}
  \end{Itemize}

\begin{figure}
\begin{center}
\includegraphics[width=0.8\linewidth]{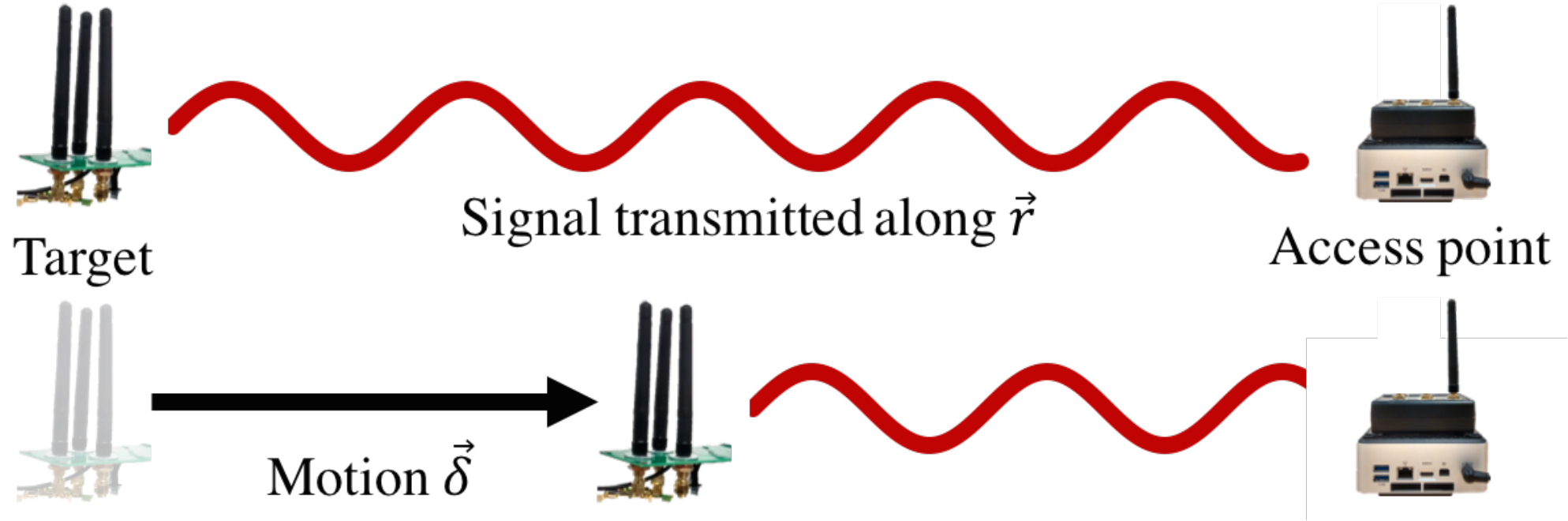}
\end{center}
\caption{The change in the phase of CSI can be modeled in terms of the displacement of the target. WiFi waves in $5$ GHz spectrum have $6$ cm wavelength. So, even millimeter-level motion creates measurable phase shift.}
\label{fig:rabbit}
\end{figure}

At a high level, as illustrated in Fig.~\ref{fig:rabbit}, \name obtains the change in the position of the transmitter by using the change in the phase of the CSI between packets. As with the ToF (Time of Flight) cameras, the phase is distorted due to the signal received from reflectors; this phenomenon is called multipath propagation. However, unlike ToF cameras where the light transmitter and camera are time-synchronized, the phase of WiFi signal is also distorted due to the lack of synchronization of clocks at the WiFi transmitter and receiver. \name tackles these challenges using novel algorithms that compensate for these distortions and provides accurate phase measurement which in turn enables accurate position tracking.

\subsection{Contributions}
\begin{Itemize}
\item \name is the \textit{first} commodity WiFi-based sub-centimeter level accurate tracking system.
\item We \textit{developed} a novel technique to overcome the distortion due to clock differences by exploiting the multipath. This is surprising as multipath is traditionally viewed as a complication in localization systems~\cite{spotfi}.
\item \name is the \textit{first} system that accurately disentangles signal from different paths by using CSI from multiple packets. The key observation is that the direction of the paths remain stationary over small intervals of time and CSI of all the packets obtained within this time can be used to resolve multipath accurately. 
%path parameters like angle of the receiver from the transmitter remain stationary over small intervals of time over which WiFi signals can provide multiple measurements which helps provide accurate estimation.
\item We \textit{built} \name using commodity Intel 5300 WiFi chips ~\cite{csitool} which demonstrated a precision of $0.25$ mm and a position tracking error of $0.88$ cm.
\end{Itemize}

\subsection{Limitations}
WiCapture's current prototype however has two limitations compared to existing systems. It has higher latency since the tracking is computed in the network and then provided as an update to the headset. Second, it is less accurate than current outside-in position tracking systems. We believe that \name's accuracy is acceptable for VR given the significant benefits that WiCapture provides around deployment, coverage and occlusion resistance.

\section{Related work}\label{sec:relatedWork}
%We classify the localization and motion tracking systems into vision-based, infrared-based, RF-based and WiFi-based systems.
\textbf{Motion tracking} has been of great interest with applications in 3D object reconstruction, virtual/augmented reality, motion capture and motion control~\cite{welch2002motion, meyer1992survey}. Systems requiring infrared LEDs or photodiodes on the tracked object have short-range, limited field of view, difficulty in tracking multiple objects, and require line of sight between tracked object and sensor~\cite{kumar2016spatial, rift, patent:20160131761, laser, prakash}. Pulsed laser light based systems, in addition, require time synchronization of multiple laser light sources~\cite{laser}. \name does not share these limitations as it has long range and typical obstructions like walls and humans are transparent to WiFi signals. 
%Also, multiple target devices can be tracked simultaneously as the signal transmitted by each WiFi card contains a hardware address unique to the WiFi card. 

Magnetic signal-based systems~\cite{raab1979magnetic,polhemus} are occlusion-resistant but have a  small range and are affected by distortions due to ferromagnetic materials ~\cite{wang1990tracking}. Radio frequency signal based techniques using RFIDs (Radio Frequency IDentification)~\cite{rfidraw, yang2014tagoram} and ultrawideband~\cite{gezici2005localization} signals demonstrated centimeter-level accuracy but have limited range and require specialized infrastructure that is not as ubiquitous as WiFi. Tracking systems using other modalities like ultrasound~\cite{foxlin1998constellation, shin2016application} and IR~\cite{vicon, dorfmuller1999robust} achieve high accuracy but require infrastructure dedicated for tracking. 

\textbf{Infrastructure-free} approaches using inertial measurement units (IMUs) can only track the orientation of a device but not the position~\cite{esser2009imu}. Visual-inertial navigation systems~\cite{vio_engel2016direct, nerurkar2014c, cadena2016past} which use cameras and IMUs track the motion of a camera using natural image features unlike earlier systems~\cite{welch1999hiball} which required instrumentation of the environment with markers. However, visual systems have problems in environments with transparent or textureless objects~\cite{patent:20160131761} and are not applicable when the camera is occluded; for example, a phone cannot track itself when placed in a user's pocket. However, when they are applicable, we view \name and infrastructure-free systems as complementary solutions that together can potentially form a robust and easy-to-deploy motion tracking system.

%Unlike the above systems, tracking an object passively without any instrumentation of the tracked object has been explored using ToF cameras~\cite{tof_doppler_heide2015doppler, shrestha2016computational, pandharkar2011estimating, ganapathi2010real}, and RADAR~\cite{cooper2008penetrating, dengler2007600, appleby2007millimeter, ralston2010real, charvat2010ultrawideband, jia2013through, xu2012novel, chetty2012through} and wireless imaging~\cite{rfcapture_adib2015capturing, joshi2015wideo, adib20143d}. However, they are limited by the size of the smallest object they can track. ~\name tracks commodity WiFi chips using commodity WiFi infrastructure providing a  ubiquitous tracking solution.

\textbf{Disentangling multipath} is a widely studied problem in Time of Flight (ToF) cameras and wireless  literature as it enables several important applications like transient light imaging ~\cite{tof_tr_o2014temporal, tof_f_lin2014fourier, tof_tr_naik2015light}, wireless imaging~\cite{rfcapture_adib2015capturing, adib20143d} and localization ~\cite{spotfi, vasisht2016decimeter}. ~\cite{tof_f_rs4010021, tof_f_freedman2014sra, tof_f_kirmani2013spumic, tof_f_droeschel2010multi, tof_f_heide2013low, tof_f_dorrington2011separating, tof_f_bhandari2014resolving, tof_f_lin2014fourier, tof_f_kadambi2016macroscopic} explored transmitting signals at multiple frequencies to resolve multipath in ToF cameras. Similarly, wireless localization systems ~\cite{spotfi, xie2016xd} explored using multiple frequencies to resolve multipath. Unlike all the previous systems, we use signal received from multiple packets to improve the accuracy of estimated multipath parameters. This hinges on the fact that the multipath parameters like direction of the propagation paths are stationary over small time periods.

\textbf{WiFi-based localization systems} can be broadly classified into  RSSI (Received Signal Strength Indicator) based, ToF based, and AoA (Angle of Arrival at the receiver) based systems. Signal strength (RSSI) based systems have 2-4 m localization error~\cite{radar,ez, Goswami2011} because of distortions in  RSSI due to multipath. RSSI fingerprint-based approaches~\cite{horus, centaur} achieve $0.6$ m error but require an expensive, recursive fingerprinting operation. ToF based approaches achieve meter-level localization error ~\cite{mariakakis2014sail, xiong2015tonetrack, vasisht2016decimeter}. AoA based approaches achieved state-of-the-art decimeter-level localization error~\cite{spotfi,arraytrack,ubicarse,cupid,pinpoint}. \name on the other hand is targeted towards estimating relative trajectory rather than the absolute position. WiFi-based tracking systems ~\cite{zeng2014your} can only classify the trajectory of the target WiFi device into limited number (four) of known gestures.

Techniques-wise, ~\cite{vanderveenJointSmoothingProof} theoretically observed  that stationarity of multipath parameters improves multipath estimation accuracy. \cite{pandharkar2011estimating} removed the effect of time offset between laser source and camera receiver on the ToF measurements by considering differential ToF between different rays. ~\name builds on these techniques and compensates not only for time offset but also frequency offset between the source and the receiver.% to provide a ubiquitous tracking system.
\section{Preliminaries}\label{sec:prel}
In WiFi systems, data is chunked into units of efficient size for communication and each unit is called as a packet~\cite{11n}. At the start of each packet, a WiFi transmitter sends a reference  signal. A WiFi receiver correlates the received signal with the reference signal and samples at time zero to calculate Channel State Information (CSI). This is similar to how image is formed in a ToF camera by correlating the received signal at each pixel with a reference signal~\cite{tof_f_bhandari2014resolving}. So, each antenna is a sensor element which performs  job similar to a pixel in a ToF camera and CSI calculated at WiFi receiver is similar to an image obtained at a ToF camera. We now provide a brief primer on CSI (or image) formation model. We included a glossary at the end of the paper for WiFi-specific terms.
\subsection{CSI calculation}
Consider the reference complex sinusoid signal  $\mathrm{e}^{j 2 \pi f t}$ of frequency $f$ emitted by the $q^\mathrm{th}$ antenna on the transmitter; here $j$ is the complex root of $-1$ and $t$ is time. The signal passes through wireless channel $h_{q}$ resulting in the received signal $h_{q} \mathrm{e}^{j 2 \pi f t}$. CSI corresponding to the $q^\mathrm{th}$ transmit antenna, $\hat{h}_{q}$, is obtained as
\vspace{-1mm}
\begin{equation}\label{eq:csiEst}
{\hat{h}}_{q} =  \frac{1}{T} \int_{0}^{T}  h_{q} \mathrm{e}^{j 2 \pi f t} \mathrm{e}^{-j 2 \pi f t + j \nu} \, dt =   h_{q} \mathrm{e}^{j \nu},
\end{equation}
where $T$ is time for which sinusoid is transmitted and $\nu$ is the phase of the receiver's local sinusoidal signal relative to the phase of the transmitter's sinusoidal signal (see Fig.~\ref{fig:clk}). 

%Consider a particular complex sinusoid signal  $\mathrm{e}^{j 2 \pi f_n t}$ corresponding to $n^\mathrm{th}$ subcarrier of frequency $f_n$ emitted by the $q^\mathrm{th}$ antenna on the transmitter; here $j$ is the complex root of $-1$ and $t$ is time. This signal passes through the wireless channel, $h_{q,n}$, gets demodulated (see Eq.~\ref{eq:csiEst}) by the local sinusoid at the receiver, and result is the CSI for that particular antenna-subcarrier pair, ${\hat{h}}_{q,n}$. Mathematically,
%\vspace{-2mm}
%\begin{equation}\label{eq:csiEst}
%{\hat{h}}_{q,n} =  \frac{1}{T} \int_{0}^{T}  h_{q,n} \mathrm{e}^{j 2 \pi f_n t} \mathrm{e}^{-j 2 \pi f_n t + j \nu} \, dt =   h_{q,n} \mathrm{e}^{j \nu},
%\end{equation}
%where $T$ is time for which sinusoid is transmitted and $\nu$ is the phase of the receiver's local sinusoidal signal relative to the phase of the transmitter's sinusoidal signal (see Fig.~\ref{fig:clk}). 

In indoor environments, similar to ToF camera systems, transmitted signal travels along multiple paths and the signals from all the paths superpose to form the received signal. Each path is associated with an AoD (Angle of Departure) from the transmitter and attenuation of the signal along the path. We now  describe the relation between the CSI and these path parameters. For this description, we consider a linear $3$-antenna array (see Fig.~\ref{fig:aoa}) although the model can be extended to any arbitrary antenna array geometry. 

\subsection{Wireless channel modeling}
\begin{figure}[t!]
\begin{center}
  \includegraphics[width=0.8\linewidth]{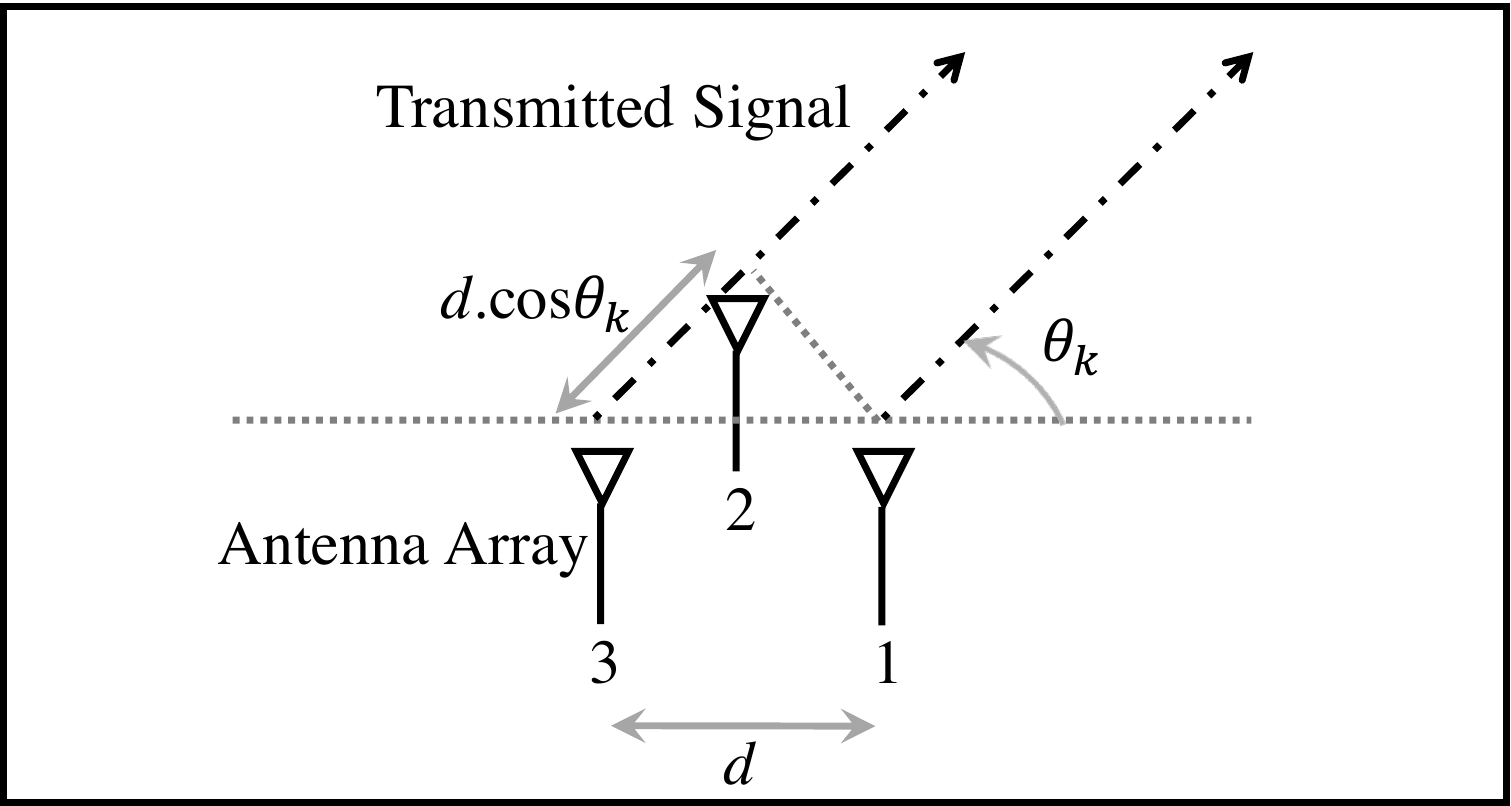}
\end{center}
  \caption{Uniform linear array consisting of $3$ antennas: For AoD of $\theta_k$, the target's signal travels an additional distance of $d \cos(\theta_k)$ from the second antenna in the array compared to the signal from the first antenna. This results in an additional phase of $-2\pi d \cos(\theta_k)/\lambda$ for the signal received from the second antenna compared to that from the first antenna.}
  \label{fig:aoa}
\end{figure}

Consider an environment with $L$ paths; \eg, the setup in Fig.~\ref{fig:arch}(b) has $2$ paths. Let the spacing between two consecutive antennas at the transmitter be $d$ (see Fig.~\ref{fig:aoa}). Let the AP/receiver be in the same plane as the transmitter antenna array. Let $\theta_k$ denote the angle of departure (AoD) from the transmitter for the $k^\mathrm{th}$
 path using the convention shown in Fig.~\ref{fig:aoa}. Let $\gamma_{k}$ denote the signal received along $k^{\mathrm{th}}$ path from the first transmitter antenna to the receiver. 

The signal along $k^\mathrm{th}$ path travels different distances from different transmit antennas to the receiver. These differences result in different phases for CSI corresponding to different antennas. As described in ~\cite{arraytrack} and illustrated in Fig.~\ref{fig:aoa}, the vector of signals received from the transmit antennas along $k^\mathrm{th}$ path can be written as $\vec{a}(\theta_k) \gamma_{k} $, where
\begin{equation}\label{eq:stvecaoa}
 \vec{a}(\theta_k) = [1     \;   \mathrm{e}^{-j 2\pi d \cos (\theta_k) /\lambda}  \; \mathrm{e}^{-j 4\pi d \cos (\theta_k) /\lambda}  ]^\top .
\end{equation}
This vector $\vec{a}(\theta_k)$ is also known as the steering vector. We have as many steering
vectors as the number of paths.

The signals from different paths superpose and result in the wireless channel $h_{q}$. So, wireless channel, 
\begin{equation}\label{eq:H1}
\textbf{H} = [\vec{a}(\theta_1) \; \ldots \; \vec{a}(\theta_L)]\textbf{F} = \textbf{A} \textbf{F},
\end{equation}
where $\textbf{H} \in \mathbb{C}^{3 \times 1}$ is the wireless channel whose $q^\mathrm{th}$ element is $h_{q}$, $\textbf{F} \in \mathbb{C}^{L \times 1}$ is the matrix of complex attenuations whose $k^\mathrm{th}$ element is $\gamma_{k}$, $\textbf{A} \in \mathbb{C}^{3 \times L}$  is the steering matrix whose $k^\mathrm{th}$ column is the steering vector for $k^\mathrm{th}$ path.

Let $\widehat{\textbf{H}} \in \mathbb{C}^{3 \times 1}$ be the observed CSI (or image) whose $q^\mathrm{th}$ element is $\hat{h}_{q}$. Wireless channel $\textbf{H}$ would be related to the observed CSI matrix using the relation, $\widehat{\textbf{H}} = \textbf{H} \mathrm{e}^{j \nu}$.

\subsection{Change in channel due to displacement}\label{pre:disp}
We will now describe how the transmitter's motion affects the wireless channel. For the rest of the paper, we will index the CSI with corresponding packet, \ie, $\widehat{\textbf{H}}_p \in \mathbb{C}^{3 \times 1}$ is CSI for $p^\mathrm{th}$ packet and ${\textbf{H}_p} \in \mathbb{C}^{3 \times 1}$ represents the corresponding wireless channel. Let's say that the transmitter moved by a small displacement $\vec{\delta}$ between the times two consecutive reference signals are transmitted. Then the path-length for the $k^\mathrm{th}$ path changes by  $(\vec{r}_{\theta_k} ^\top \cdot \vec{\delta})$ where $\vec{r}_{\theta_k}$ is the unit vector along the direction of departure for the particular path with AoD $\theta_k$. This induces a phase shift, $2\pi (\vec{r}_{\theta_k}^\top \cdot \vec{\delta})/\lambda$. So, attenuation of the $k^\mathrm{th}$ path gets multiplied by $\mathrm{e}^{-j2\pi (\vec{r}_{\theta_{k}}^\top \cdot \vec{\delta})/\lambda}$. Mathematically, if $\textbf{H}_1 = \textbf{A} \textbf{F}$ as in Eq.~\ref{eq:H1}, then the wireless channel for the second packet is 
\begin{equation}\label{eq:H2}
\textbf{H}_2 = \textbf{A}  \textbf{D} \textbf{F},
\end{equation}
where ${\textbf{H}_2} \in \mathbb{C}^{3 \times 1}$, $\textbf{A} \in \mathbb{C}^{3 \times L}$, ${\textbf{F}} \in \mathbb{C}^{L \times 1}$, and $\textbf{D} \in \mathbb{C}^{L \times L}$ is a diagonal matrix with entries $\mathrm{e}^{-j2\pi (\vec{r}_{\theta_1}^\top \cdot \vec{\delta})/\lambda}$, $\ldots$, $\mathrm{e}^{-j2\pi (\vec{r}_{\theta_{L-1}}^\top \cdot \vec{\delta})/\lambda}$, and  $\mathrm{e}^{-j2\pi (\vec{r}_{\theta_L}^\top \cdot \vec{\delta})/\lambda}$.

\subsection{Phase distortion due to frequency offset}\label{pre:cfo}
\begin{figure}[t!]
  \includegraphics[width=1\linewidth]{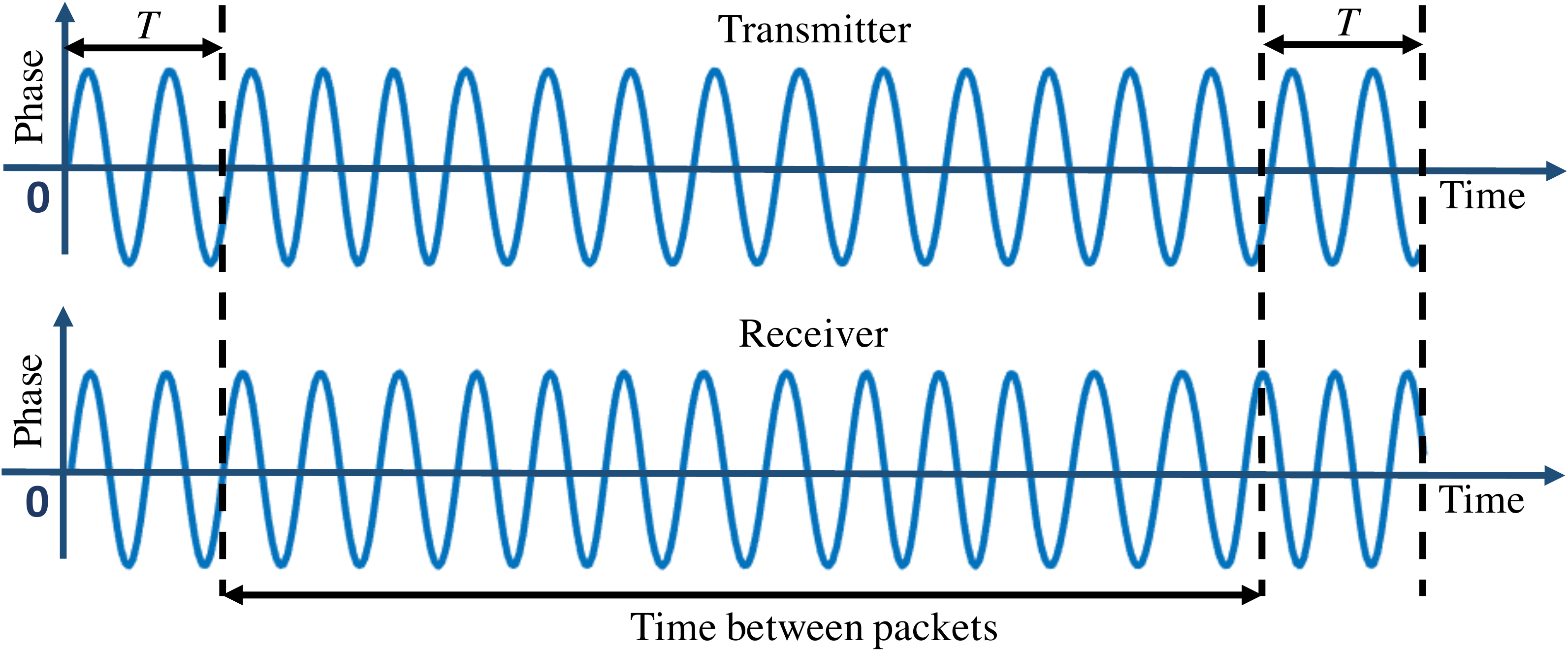}
  \caption{$T$ in Eq.~\ref{eq:csiEst} is on the order of few microseconds which is very small compared to the time interval between successive WiFi transmissions which is on the order of milliseconds. So, the relative phase can be assumed to be constant for $T$ period but changes on the order of few hundred radians within few milliseconds.}
  \label{fig:clk}
\end{figure}

The transmitter and the receiver WiFi chips have different clocks and this creates an offset between the frequencies of the local reference sinusoids used at the transmitter and the receiver. Moreover, the offset is not constant with time as the clocks drift~\cite{zucca2005clock, brown2012fundamental}. As illustrated in Fig.~\ref{fig:clk}, the frequency offset results in a change in the relative phase between the sinudoids at the transmitter and the receiver (see Equation \ref{eq:csiEst}) from packet to packet. Let $\nu_p$ be the relative phase between the two sinusoids for $p^\mathrm{th}$ packet. Then, 
\begin{equation}\label{eq:Hp}
\widehat{\textbf{H}}_p = {\textbf{H}_p} \mathrm{e}^{j \nu_p}. 
\end{equation}
So, if the transmitter moved by $\vec{\delta}$ between consecutive packets, then using Equations~\ref{eq:H1},~\ref{eq:H2} and ~\ref{eq:Hp}, the CSI reported by the WiFi chip for the two packets can be written as 
\begin{equation}\label{eq:H2_hat}
\widehat{\textbf{H}}_1 = \textbf{A} \textbf{F} \mathrm{e}^{j \nu_1}, \widehat{\textbf{H}}_2 = \textbf{A}  \textbf{D} \textbf{F} \mathrm{e}^{j \nu_2}.
\end{equation}

Eq.~\ref{eq:H2_hat} relates the observed CSI to the transmitter's displacement. It is important to note that frequency offset of $20$ kHz is typical and WiFi standard allows it to be as high as $200$ kHz~\cite{11n}. This implies that the distortion in CSI due to frequency offset is orders of magnitude greater than the change caused by transmitter's motion. For example, if two consecutive WiFi packets are sent with a time-gap of $10$ ms (which is typical), then the relative phase between the transmitter and receiver sinusoids can change by $630$ radians and is observed in practice~\cite{nandakumar2014wi}. To compare, if the transmitter moved by $5$ mm during the same $10$ ms, then the change in the phase of any path is less than $0.6$ radians.

\section{Design}
\begin{figure*}
\begin{center}
\includegraphics[width=1\linewidth]{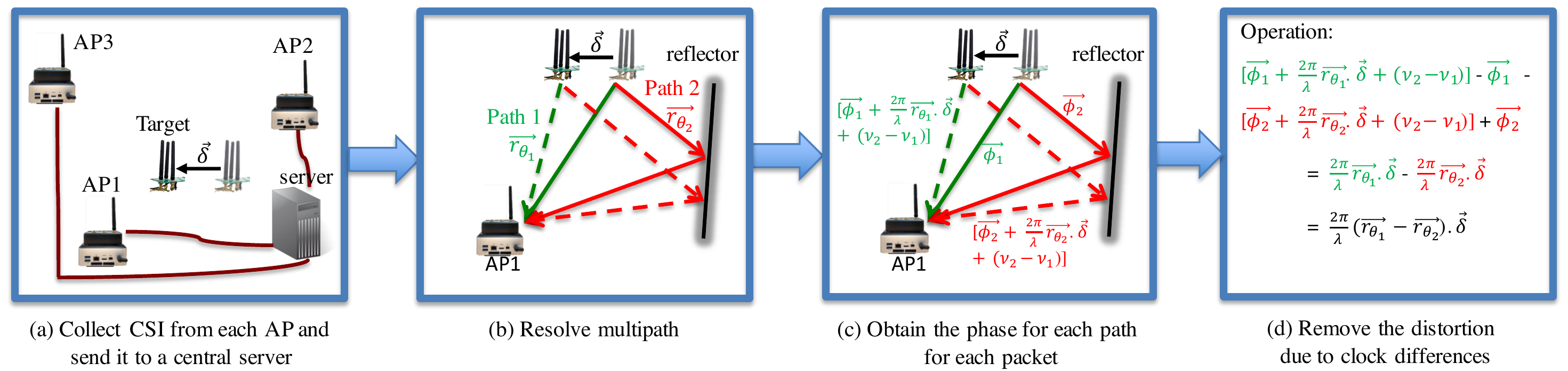}
\end{center}
\caption{Target moves by $\vec{\delta}$. The new position of the target is shown with solid target icon and the old position with shaded icon. In the first step, for each AP, \name obtains unit vectors along the direction of departure of all the paths. Here, there are two paths, green and red paths, with directions $\vec{r}_{\theta_1}$ and $\vec{r}_{\theta_2}$. The paths for first packet are shown using solid lines and the paths for the second packet using dashed lines. Next, the phase of complex attenuation along both the paths for both the packets is calculated. Finally, the effect of frequency offset on the phase values, $\nu_2 - \nu_1$, is removed to obtain equations dependent on just $\vec{r}_{\theta_1}$, $\vec{r}_{\theta_2}$ and $\vec{\delta}$, which can be solved to estimate $\vec{\delta}$.}
\label{fig:arch}
\end{figure*}
% $\protect\overrightarrow{r_{\theta_1}}$ and $\protect\overrightarrow{r_{\theta_2}}$

Fig.~\ref{fig:arch} shows how \name would be deployed and summarizes the solution procedure. The target transmits normal WiFi signals. Each WiFi AP/receiver calculates CSI (or image) and sends it to a central server. The server uses CSI from all the access points (APs) to determine the position
of the target by performing the following two steps:
\vspace{-4mm}
\begin{Enumerate}
\item Estimate AoD and complex attenuation of all the paths.
\item Use the attenuation from consecutive CSIs along with AoD of the paths to estimate the displacement of the transmitter during the time between consecutive CSIs.
\end{Enumerate}

\subsection{Estimating AoD of all the paths}\label{sec:pathEst}
Let us rewrite the CSI for the first packet from Eq.~\ref{eq:H2_hat} as $\widehat{\textbf{H}}_1 = \textbf{A} (\textbf{F} \mathrm{e}^{j \nu_1})$. We can observe that CSI is nothing but a linear combination of steering vectors in $\textbf{A}$. If one obtains the steering vectors, then finding the AoD of all the paths is trivial using Eq.~\ref{eq:stvecaoa}. However, the problem of obtaining steering vectors from a single linear combination of theirs is ill-posed. Prior work~\cite{vanderveenJointSmoothingProof} has theoretically investigated that the more number of linear combinations of steering vectors with independent weights are provided, the higher the accuracy of the estimated steering vectors. 

Note that the AoDs of paths are relatively stationary and do not change rapidly from packet to packet. For example, if the receiver is $3$ m away from the transmitter, a tiny motion of $1$ mm changes the AoD of any path by atmost $0.02$ degrees, which does not cause any measurable changes in the elements of the steering matrix. However, the weights in the linear combination change significantly as even a small motion creates significant change in the phase of complex attenuation along a path. 

Using this insight, CSI from $P$ packets where AoD has not changed much are concatenated using Eq.~\ref{eq:csiCon}. CSI for any of these packets is linear combination of the same steering vectors. So,
\begin{equation}\label{eq:csiCon}
\textbf{X} = [\widehat{\textbf{H}}_1 \; \widehat{\textbf{H}}_2 \; \ldots \; \widehat{\textbf{H}}_P]
 = \textbf{A} \textbf{G},
\end{equation}
where $\textbf{X} \in \mathbb{C}^{3 \times P}$, $\textbf{A} \in \mathbb{C}^{3 \times L}$ is steering matrix which is common for all the $P$ packets, $\textbf{G} \in \mathbb{C}^{L \times P }$ is a matrix of steering vector weights. This is standard form for applying well-known MUSIC algorithm to compute AoD~\cite{music, spotfi, xie2016xd}. The goal of the MUSIC algorithm is to find all the steering vectors given multiple linear combinations of the steering vectors and it provides an accurate estimate of $\textbf{A}$ due to diverse linear combinations.

%We omit the mathematical
%derivation for brevity, but we can show the following: \emph{the eigenvectors corresponding to zero eigenvalues of} ${\widehat{\textbf{H}}_1}{\widehat{\textbf{H}}_1}^{*}$ \emph{are orthogonal to the steering vectors in} ${\textbf{A}}$\footnote{$ \textbf{B}^{*} $ is conjugate transpose of $ \textbf{B} $.}. So the MUSIC algorithm
%at a basic level proceeds first by computing $\widehat{\textbf{H}}_1 \widehat{\textbf{H}}_1^{*}$, then computing the
%eigenvectors corresponding to its zero eigenvalues, and then
%finding the steering vectors orthogonal to these vectors. 

\subsection{Estimating the displacement of the transmitter between consecutive packets}\label{sec:disp}
We will now show how the displacement of the WiFi transmitter between two consecutive packets can be estimated. Consider the CSI from the first two packets.

\noindent\textbf{Estimate the steering vectors' weights: }
CSI is a linear combination of steering vectors estimated using procedure in Sec.~\ref{sec:pathEst}. For $p^{\mathrm{th}}$ packet, these weights in the linear combination, $\widehat{\textbf{F}}_p \in \mathbb{C}^{L \times 1}$, can be obtained using 
\begin{equation}\label{eq:AhatH}
\widehat{\textbf{F}}_p = \textbf{A}^\dagger \widehat{\textbf{H}}_p. \footnote{$ \textbf{B}^\dagger $ is the pseudo-inverse of $ \textbf{B} $.}
\end{equation}
Substituting Eq.~\ref{eq:H2_hat} into Eq.~\ref{eq:AhatH}, we can observe that 
\begin{equation}\label{eq:Fhat}
\widehat{\textbf{F}}_1 = \textbf{F} \mathrm{e}^{j \nu_1}, \widehat{\textbf{F}}_2 = \textbf{D} \textbf{F} \mathrm{e}^{j \nu_2}
\end{equation}
%Note that $\textbf{D} \in \mathbb{C}^{L \times L}$ and it's entries contain information about target's displacement (see Eq.~\ref{eq:H2}). 

\noindent\textbf{Estimate the change in the complex attenuation for each of the paths: }
Note that $\textbf{D} \in \mathbb{C}^{L \times L}$ and it's entries contain information about target's displacement (see Eq.~\ref{eq:H2}). We obtain an estimate ${\textbf{D}}$ by solving the convex optimization problem \ref{eq:con1} using standard procedures~\cite{ConvexBoyd}.
\begin{equation}\label{eq:con1}
\begin{aligned}
& \underset{{\textbf{D}}}{\text{minimize}}
& &  \|\widehat{\textbf{F}}_2 - {\textbf{D}} \widehat{\textbf{F}}_1 \| \\
& \text{subject to}
& & {\textbf{D}} \text{ is diagonal}.
\end{aligned}
\end{equation}
From Equations~\ref{eq:Fhat},~\ref{eq:con1} and~\ref{eq:H2}, the $k^\mathrm{th}$ diagonal element of ${\textbf{D}}$, ${\textbf{D}}_{k,k}$, is an estimate of $\mathrm{e}^{-j2\pi (\vec{r}_{\theta_k}^\top \cdot \vec{\delta})/\lambda + j \nu_2 - j\nu_1}$. Note that unit vector in the direction of the departure can be obtained using $\vec{r}_{{\theta}_k} = [\cos({\theta}_k) \; \sin({\theta}_k)]^\top$; here ${\theta}_k$ is the AoD for the particular path estimated using procedure in Sec.~\ref{sec:pathEst}. So, if the term $(\nu_2 - \nu_1)$ is removed, one can estimate $\vec{\delta}$ from the phase of elements of $\textbf{D}$. 

However, as discussed in Sec.~\ref{pre:cfo}, the term $\nu_2 - \nu_1$ due to frequency offset is orders of magnitude larger than the term $-2\pi (\vec{r}_{{\theta}_k}^\top \cdot \vec{\delta})/\lambda$ due to displacement. This change in phase due to frequency offset is not only unavailable but also extremely hard to predict~\cite{brown2012fundamental} making it hard to estimate and remove the term $\nu_2 - \nu_1$. This is precisely the phenomenon that led to long-held notion that phase information across packets is uncorrelated in commodity WiFi systems~\cite{nandakumar2014wi} and is unusable for  tracking. 

%This is also precisely the reason why radar-based motion tracking systems have never been adapted to WiFi systems so far. In radar systems, the transmitter and the receiver are the same. So the clocks are sychronized by definition and there is no frequency offset. Then, the relative phase between the target and the AP does not change. Hence, $\nu_2 = \nu_1$. The phase of $\widehat{\textbf{D}}_{k,k}$ can be used to obtain the target's displacement if the AoD information is provided/estimated. 

\noindent\textbf{Use attenuation-change between paths: }%\label{sec:solveDisp}
Our unique insight is that one can get rid of the effect of frequency offset by using the phase of the signal from multiple paths. This is surprising as multipath is traditionally viewed as a complicating distortion that needs to be compensated in many ToF depth sensor and WiFi localization systems~\cite{spotfi, tof_sparse_kadambi2013coded}. Notice that the effect of frequency offset is the same for all the paths, \textit{i.e.}, the term, $\nu_2 - \nu_1$, is present in the change of phase for all the paths. So, effect of the clock differences is removed by considering the change in the phase of signal along a path with respect to change in the phase of signal from another path.

Specifically, consider the  phase of the complex number ${\textbf{D}}_{k+1,k+1}/{\textbf{D}}_{1,1}$. It is an estimate of $(-2\pi \vec{r}_{{\theta}_{k+1}}^\top \cdot \vec{\delta}/\lambda + \nu_2 - \nu_1) - (-2\pi \vec{r}_{{\theta}_1}^\top \cdot \vec{\delta}/\lambda + \nu_2 - \nu_1) = (-2\pi (\vec{r}_{{\theta}_{k+1}} - \vec{r}_{{\theta}_1})^\top \cdot \vec{\delta}/\lambda )$. Notice that the term, $\nu_2 - \nu_1$, canceled out by performing this operation. So, a $(L-1)$-dimensional vector $\vec{s}$ is calculated whose $k^{\mathrm{th}}$ element is 
\begin{equation}\label{eq:s}
\vec{s}_k = \text{phase of } ({\textbf{D}}_{k+1,k+1}/{\textbf{D}}_{1,1}).
\end{equation}
Then $\textbf{R} \in \mathbb{R}^{(L-1) \times 2}$ is calculated whose $k^\mathrm{th}$ row is 
\begin{equation}\label{eq:R}
\frac{-2\pi}{\lambda} [(\textrm{cos}({{{\theta}}_{k+1}})-\textrm{cos}({{{\theta}}_{1}})) \;\;  (\textrm{sin}({{{\theta}}_{k+1}})-\textrm{sin}({{{\theta}}_{1}}))].
\end{equation}
%$(-2\pi* (\vec{r}_{{\theta}_{k+1}} - \vec{r}_{{\theta}_1})/\lambda$ 
One can then obtain an estimate of the displacement by solving the simple linear least squares problem \ref{eq:con2}.
\begin{equation}\label{eq:con2}
\vec{{\delta}} = \arg\!\min \| \textbf{R} \vec{{\delta}} - \vec{s}  \|. 
\end{equation}

If there are multiple APs, matrix $\textbf{R}$ and vector $\vec{s}$ obtained from multiple access points are concatenated vertically. If there are $L$ paths from the target to each of the $U$ APs, then the concatenated $\textbf{R} \in \mathbb{R}^{U*(L-1) \times 2}$ and the concatenated $\vec{s} \in \mathbb{R}^{U*(L-1)}$. Since the target displacement is same irrespective of the AP, one can estimate of the displacement by solving Eq.~\ref{eq:con2} using these concatenated matrices. We summarize the overall algorithm in Algorithm \ref{algo:mt}. 

\vspace{-3mm}
 \begin{algorithm}
 \caption{\name 's motion tracking algorithm}
 \label{algo:mt}
 \KwData{CSI of packets from target to each of the $U$ APs}
 \KwResult{Trajectory traced by the target }
  Initiate the trajectory at origin \;
  \For{\text{each packet $p$ received at APs} }{
   Consider packets received within the last $V$ seconds. Let the number of such packets be $P$ \;
   Form $\textbf{X}$ from CSI of $P$ packets using Eq.~\ref{eq:csiCon} \;
%   Compute $E_S$ whose columns are top $L$ eigenvalues \;
%   Evaluate MUSIC spectrum $P_{MU}(\theta) = {\left( \vec{a}^\mathrm{H}(\theta) E^{*}_S E_S  \vec{a}(\theta) \right)} $ \;
 %  Obtain AoD of multipath components as peaks of $P_{MU}(\theta)$ \;
Apply MUSIC~\cite{music} on $\textbf{X}$ to find AoD of $L$ paths 	\;
	  Obtain the steering vector weights using Eq.~\ref{eq:AhatH} \;
	  Obtain change in complex attenuation between $p^{\mathrm{th}}$ and $(p+1)^{\mathrm{th}}$ packets by solving~\ref{eq:con1} \;
	  Form $\textbf{R}$ using Eq.~\ref{eq:R} and $\vec{s}$ using Eq.~\ref{eq:s} \;
	  Update the trajectory by adding displacement obtained by solving~\ref{eq:con2} \; 
	  }
\end{algorithm}
\vspace{-4mm}

%\begin{equation}\label{eq:H2_hat}
%\widehat{\textbf{H}}_1 = \textbf{A} \textbf{F} \mathrm{e}^{j \nu_1}, \widehat{\textbf{H}}_2 = \textbf{A}  \textbf{D} \textbf{F} \mathrm{e}^{j \nu_2}.
%\end{equation}

\section{Evaluation}
We implemented \name using off-the-shelf Intel 5300
WiFi cards which support three antennas. We employed Linux CSI tool~\cite{csitool} to obtain the
CSI. The WiFi cards operate in 5 GHz WiFi spectrum ($f$ in Sec.~\ref{sec:prel}). Also, the CSI information is quantized,
i.e., each of real and imaginary parts of CSI is represented using 8 bits.

The system used for evaluation consists of APs and a target device equipped with WiFi cards. The target/transmitter has a $3-$antenna circular array with distance between any two antennas equal to $2.6$ cm. The APs operate in monitor mode. CSI is calculated once every $6$ ms. So, the trajectory is estimated at an update rate of $167$ Hz for our evaluation experiments. An implementation is provided in~\cite{kotaru_wic_code}. We use $U=4$ APs, set $V=10$ s and $L=2$ paths in Algorithm~\ref{algo:mt}. 

\subsection{Stationary experiments}
We start by examining the jitter/precision of \name, \ie, how stationary the estimated position is when the target is stationary. This is important for VR because the scene displayed on the VR headset is not expected to change when the user is not moving. In our experiment, the target remains stationary and transmits $1000$ packets. The access points are placed at the corners of $5$ m $\times$ $6$ m space. The standard deviation of the trajectory obtained by using Algorithm~\ref{algo:mt} is used as the measure of jitter. The median jitter observed over 21 experiments at different target positions is $0.25$ mm.

We measured the jitter of Oculus DK2 system by placing it at the same positions. The Oculus camera is placed at $0.75$ m away from the headset. The median jitter observed for Oculus DK2 is $0.10$ mm. So, \name's jitter in position estimation is comparable to that of a commercial position tracking system which requires dedicated infrastructure.

\subsection{Controlled tracking experiments}\label{sec:controlled}
\begin{figure}[t!]
\hfill
\subfigure[Trajectory 1]{\includegraphics[width = 0.49\linewidth]{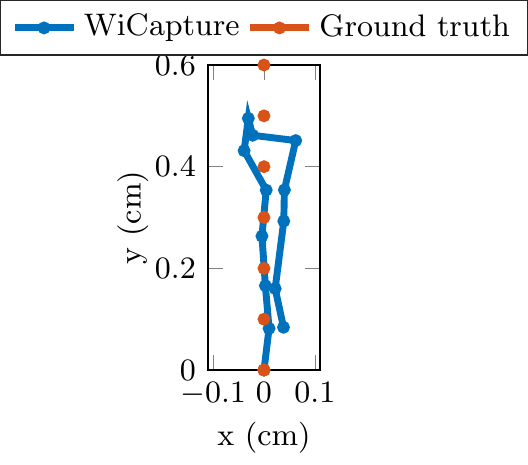}}
\hfill
\subfigure[Trajectory 2]{\includegraphics[width = 0.49\linewidth]{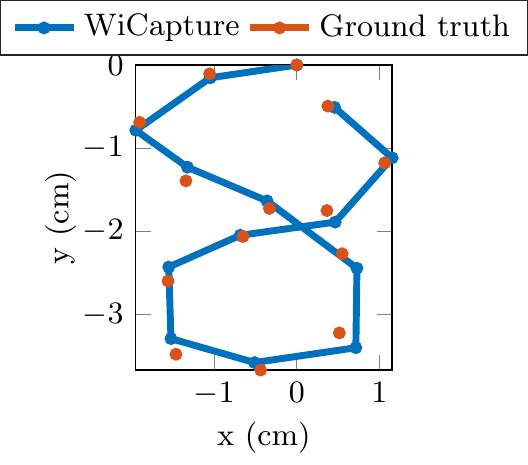}}
\caption{Controlled trajectories where the target is positioned at different locations in a trajectory mechanically.}
\label{fig:mmStage}
\end{figure}

Next we evaluate the resolution of \name. We mount the target on a mechanical stage which has a least count of $0.005$ cm. The target is moved in increments of $0.1$ cm and then in decrements of $0.1$ cm so that it reaches the initial position as shown in Fig.~\ref{fig:mmStage}(a). The target transmits a WiFi packet at each position. Trajectory estimated by \name is translated so that the initial position is origin. The maximum error in estimation of position of any point in the trajectory is $0.11$ cm.

We conducted another experiment where the target is moved mechanically to different positions on a trajectory shown in Fig.~\ref{fig:mmStage}(b). The maximum error in estimation of position of any point in the trajectory is $0.27$ cm. Note that \name resolves even millimeter-level target motion.

\subsection{Extensive tracking experiments}\label{sec:extensive}
Motion tracking accuracy of \name is dependent on the multipath environment,
the material used in walls, the presence of metallic
objects, the density of WiFi AP deployment and many other
factors. In this evaluation, we test \name's accuracy in different deployment scenarios.

\begin{figure*}[!htb]
\hfill
\subfigure[Indoor office deployment]{\includegraphics[width = 0.32\linewidth]{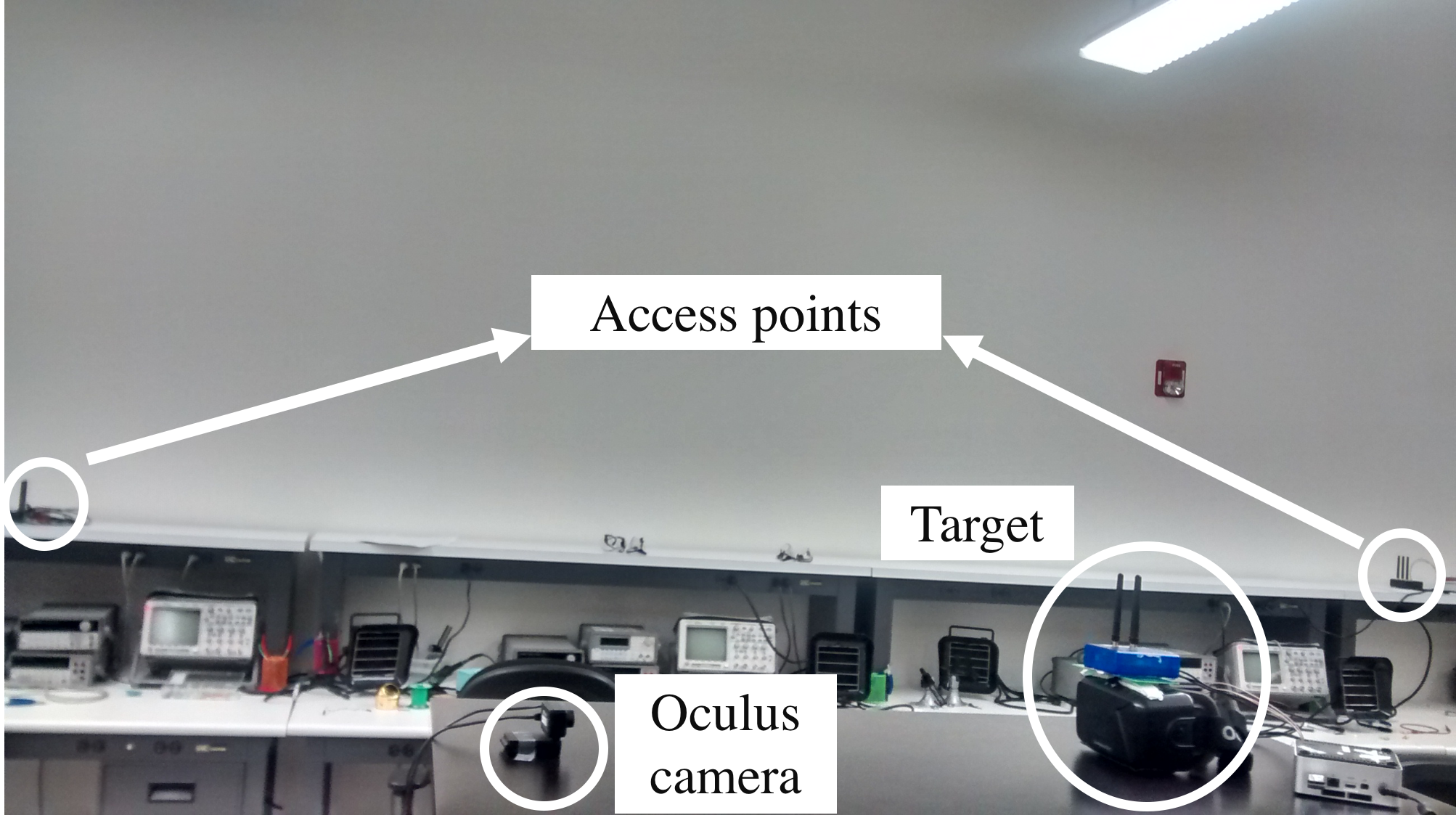}}
\hfill
\subfigure[Occlusion deployment]{\includegraphics[width = 0.32\linewidth]{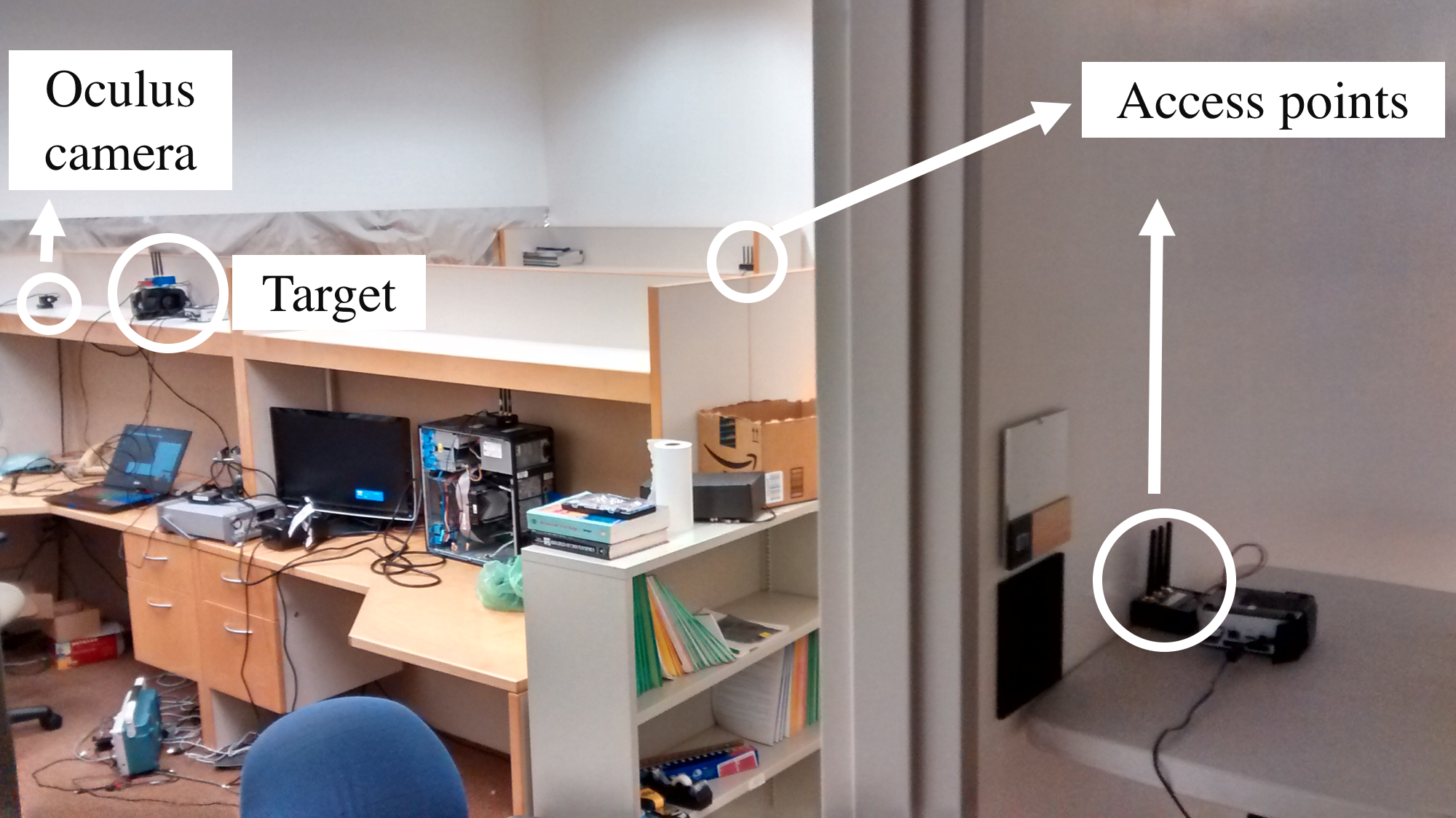}}
\hfill
\subfigure[Outdoor deployment]{\includegraphics[width = 0.32\linewidth]{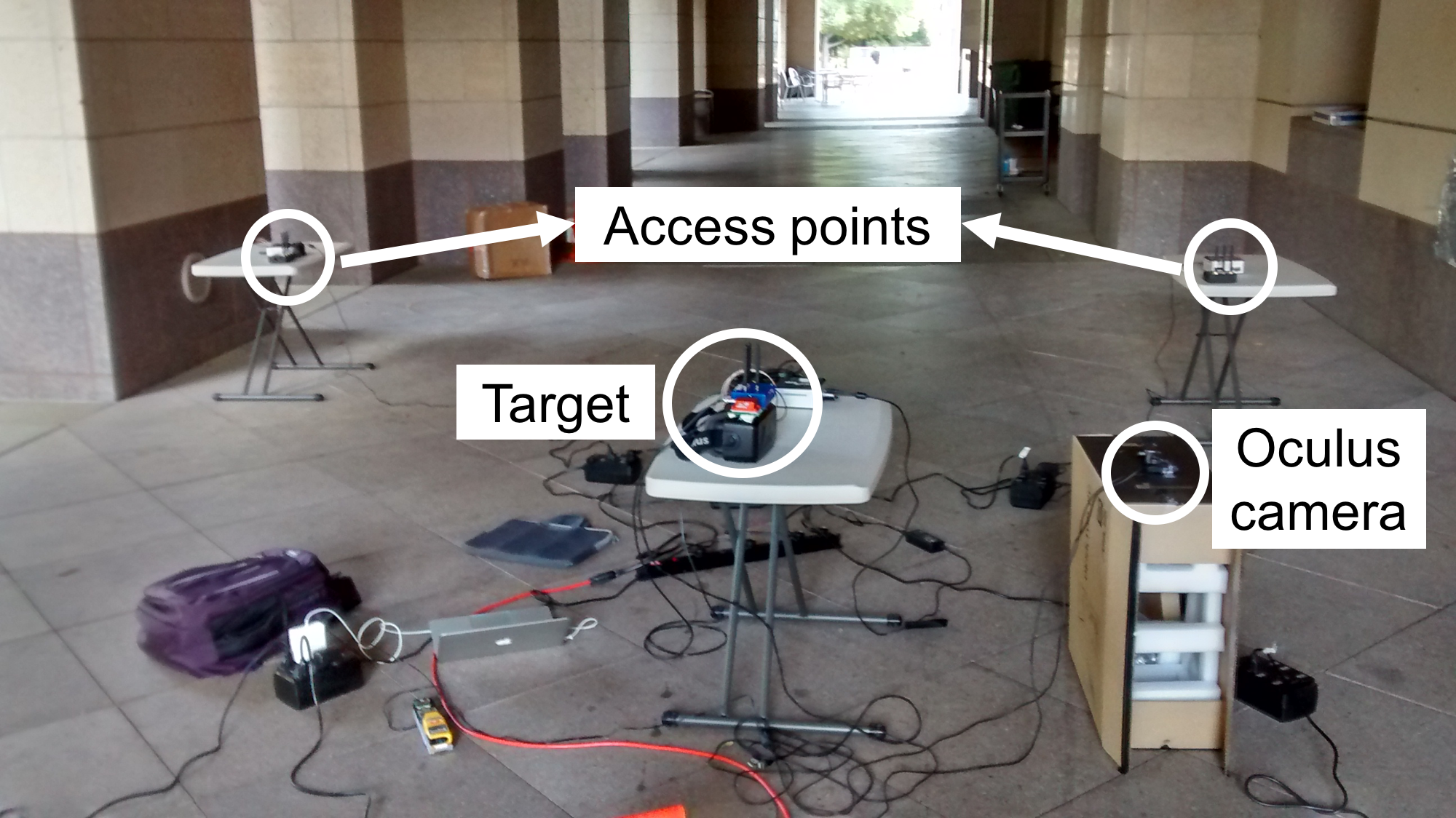}}
\caption{Experiment setups for the three deployments}
\label{setup}
\end{figure*}
%\begin{figure*}[!htb]
%\hfill
%\subfigure[Indoor office deployment]{\includegraphics[width = 0.32\linewidth]{./figEval/losTraj}}
%\hfill
%\subfigure[Occlusion deployment]{\includegraphics[width = 0.32\linewidth]{./figEval/nlosTraj}}
%\hfill
%\subfigure[Outdoor deployment]{\includegraphics[width = 0.32\linewidth]{./figEval/outsideTraj}}
%\caption{Sample traces for the three deployments}
%\label{traj}
%\end{figure*}
\begin{figure*}[!htb]
\hfill
\subfigure[Indoor office deployment]{\includegraphics[width = 0.32\linewidth]{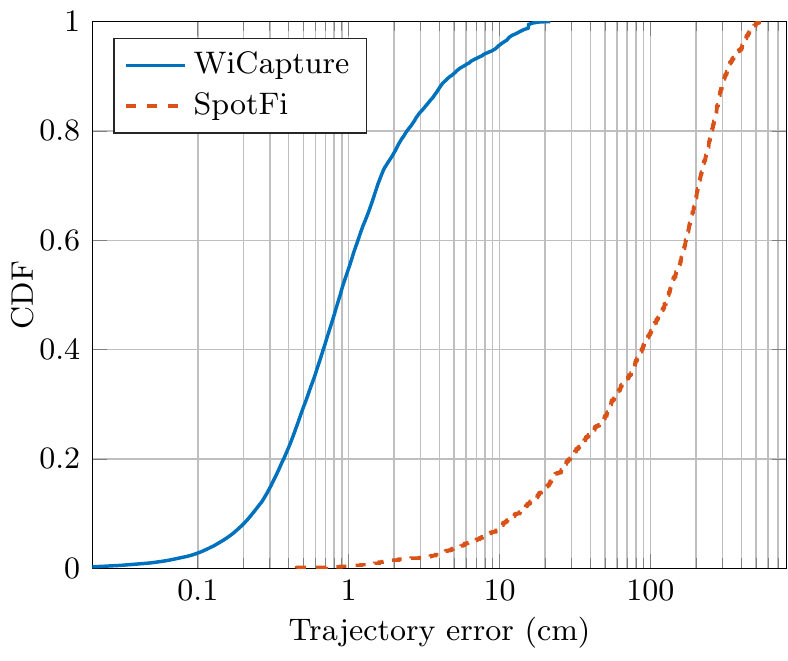}}
\hfill
\subfigure[Occlusion deployment]{\includegraphics[width = 0.32\linewidth]{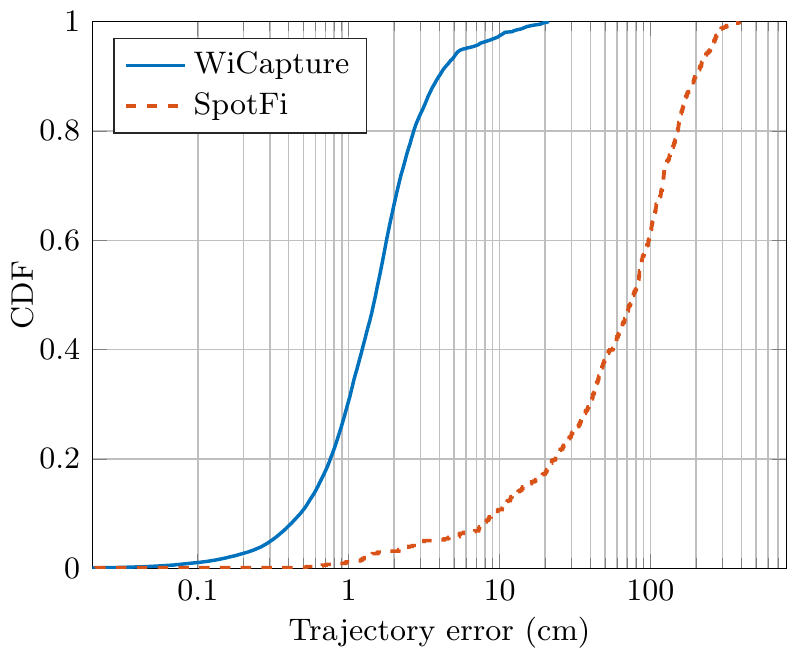}}
\hfill
\subfigure[Outdoor deployment]{\includegraphics[width = 0.32\linewidth]{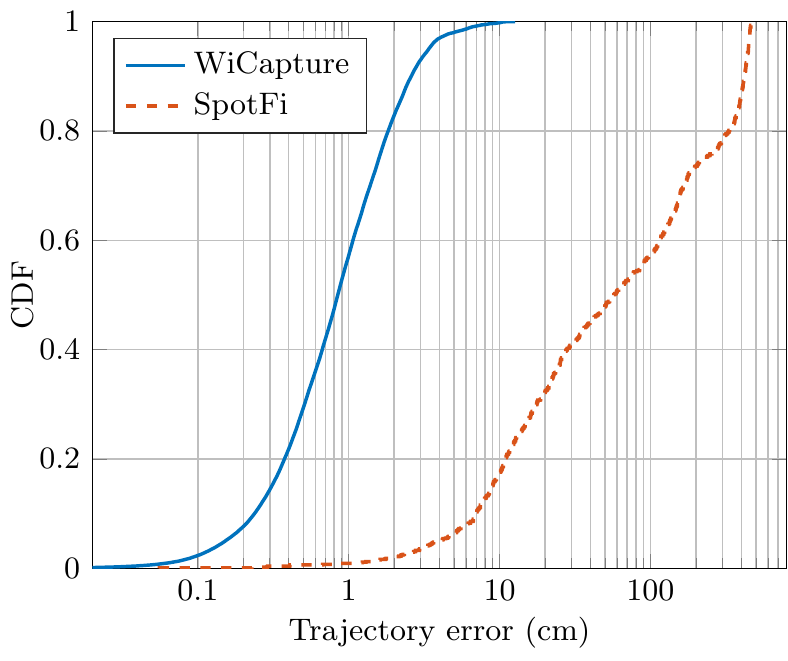}}
\caption{Cumulative Distribution Function (CDF) of trajectory error for the three deployments}
\label{err}
\end{figure*}

\noindent\textbf{Compared Approaches:} To the best of our knowledge, there exists no WiFi-based system that is geared towards \textit{tracking} the motion of a commodity WiFi chip. We faithfully implemented SpotFi, a state-of-the-art WiFi localization system and compared \name against it. We did not compare with other non-vision based systems like ultrasound~\cite{liu2013guoguo} because of limitations detailed in section~\ref{sec:relatedWork} such as extensive installation of dedicated infrastructure.

\noindent\textbf{Ground truth: }We use an Oculus DK2~\cite{dk2Video} headset which is rigidly attached to the target to obtain ground truth trajectory. We placed the headset $0.9$ m away from the Oculus camera. Under these conditions, we calculated the accuracy of Oculus position tracking to be at sub-millimeter level using mechanical stage experiments (see Sec.~\ref{sec:controlled}).

\noindent\textbf{Metric (Trajectory error): }To focus on the shape of the trajectory alone, we translate the trajectory reported by each of the systems (Oculus DK2, SpotFi and \name) by the initial position of the trajectory so that the initial position of all the trajectories is origin. Similarly, since the reference coordinate axes of different systems are not aligned, we rotate the trajectory reported by \name (and SpotFi) so that the Root Mean Squared Error (RMSE) between the points of the rotated trajectory of \name (and SpotFi) and the points on the Oculus trajectory is minimized. Note that we just shift and rotate the trajectories of \name and SpotFi but do not scale the trajectories. As in other motion tracking systems~\cite{rfidraw}, absolute point-by-point position difference between this shifted and rotated trajectories and the ground truth trajectory is reported as the trajectory error.

\substart
\subsubsection{Indoor office deployment}\label{sec:los}
\subend
\textbf{Method:} We deployed \name in a $5$ m $\times$ $6$ m room with access points placed at the four corners and the target is in the same room. This is the typical access point deployment density used in state-of-the-art WiFi localization systems~\cite{spotfi}. We traced $97$ trajectories with the target device. Fig.~\ref{fig:trace} shows a sample trajectory. The target was moved in a continuous manner on a table rather than point by point as done in Sec.~\ref{sec:controlled}. Fig.~\ref{setup}(a) shows the experimental setup.

\noindent\textbf{Analysis: }From Fig.~\ref{err}(a), \name achieves a median trajectory error of $0.88$ cm. SpotFi's median trajectory error, $132$ cm, is more than two orders of magnitude larger than that of \name. \name thus achieves sub-centimeter-level motion tracking using commodity WiFi.
\substart
\subsubsection{Occlusion deployment}
\subend
\textbf{Method:} We evaluate \name under challenging conditions where the target is one room and all the APs are occluded from the target either by furniture or by walls. Fig.~\ref{setup}(b) shows the experimental setup which shows couple of APs where one AP is placed outside the room where the target is placed and another AP is separated from the target through a cubicle. We traced $64$ different trajectories.

\noindent\textbf{Analysis: }Fig.~\ref{err}(b) plots the CDF of the trajectory error of \name and SpotFi when either one or two access points are outside the room where the target is placed. However, note that all the access points are occluded whether they are inside the target's room or not. Under these conditions, \name achieves a median trajectory error of $1.51$ cm. Thus, even in the challenging conditions where the target is occluded from the positioning system, \name achieves accuracy acceptable for many applications.

We further stress-tested by placing three or more APs outside the room at which point the reconstructed trajectories accumulated large errors. So, \name can provide accurate tracking only when atleast two access points are present which are not separated from the target by walls.
\substart
\subsubsection{Outdoor deployment}
\subend
\textbf{Method:} We evaluate \name in a $5$ m $\times$ $6$ m outdoor space. The experiments are conducted in shade for accurate Oculus measurements. Fig.~\ref{setup}(c) shows the experimental setup. We traced $89$ trajectories.

\noindent\textbf{Analysis: }From Fig.~\ref{err}(c), \name achieves a median trajectory error of $0.85$ cm compared to SpotFi's $57$ cm trajectory error. Thus, \name can ubiquitously track a commodity WiFi device as long as there is WiFi infrastructure irrespective of whether the target is indoors or outdoors.

\subsection{Deep dive into \name}
\name achieves accurate motion tracking due to two novel techniques. First, \name accurately resolves multipath by using multiple packets for estimation. Second, \name accurately removes the phase distortion due to frequency offset between the transmitter and the receiver. We now test the significance of each of these factors individually. For the following experiments, we consider the data from the indoor office deployment scenario in Sec.~\ref{sec:los}.
\begin{figure}[t!]
\hfill
\subfigure[Improved AoD estimation accuracy]{\includegraphics[width = 0.49\linewidth]{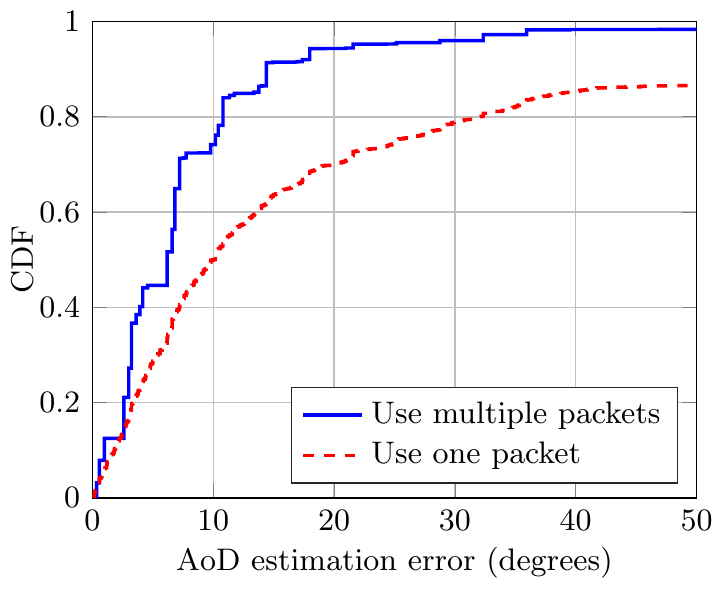}}
\hfill
\subfigure[Removing phase distortion due to different clocks]{\includegraphics[width = 0.49\linewidth]{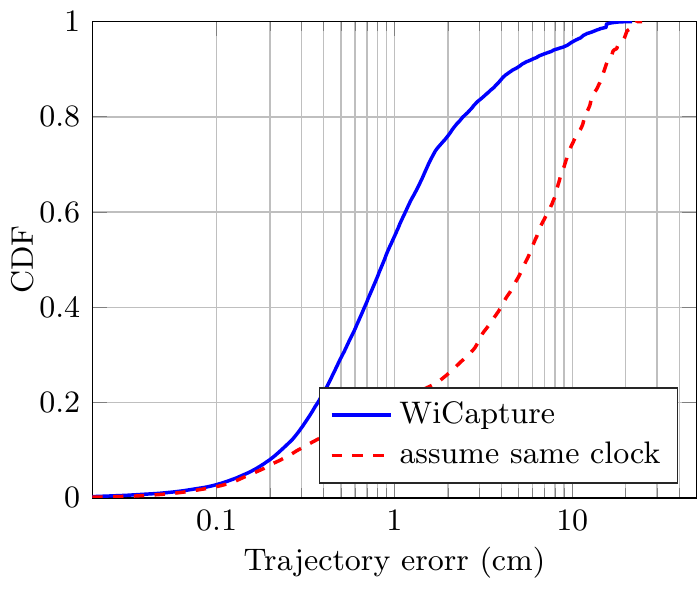}}
\caption{(a) plots the CDF of AoD estimation error when using multiple packets (\name) and when using one packet. (b) plots the CDF of trajectory error when using \name and when using \textit{assume same clock} method which ignores frequency offset.}
\label{deepdive}
\end{figure}
\substart
\subsubsection{Improved AoD estimation}
\subend
\textbf{Method:} We compare \name against an alternate method which uses single packet for AoD estimation (set $P$ equal to $1$ in Eq.~\ref{eq:csiCon}). Error of a particular method (single packet or multiple packets) is measured by using the absolute difference between ground truth direct path AoD (which is measured manually) and AoD estimated using the particular method that is closest to
this ground truth.

\noindent\textbf{Analysis:} From Fig.~\ref{deepdive}(a), $80^{\mathrm{th}}$ percentile AoD estimation error by using multiple packets is $11$ degrees and is $3\times$ smaller than the error obtained by using a single packet.
\substart
\subsubsection{Removing phase distortion from frequency offset}
\subend
\textbf{Method:} We compare \name against an alternate method which solves a system of equations where each equation equates the change in the phase of complex attenuation of a path directly as a linear function of displacement without considering the offset from the clock differences. We call this alternate method as \textit{assume same clock}.

\noindent\textbf{Analysis:} From Fig.~\ref{deepdive}(b), the trajectory estimation errors are $6 \times$ worse when frequency offset is ignored.

\section{Discussion and conclusion}
All the major WiFi chip families (Atheros, Intel, and Marvell) expose CSI~\cite{csitool, cupid}. Hence, we believe \name can be added to any commodity WiFi infrastructure.

To enable high-speed communication, WiFi chips transmit reference signals on multiple frequencies and receivers, typically, have multiple antennas. Algorithm~\ref{algo:mt} can be extended to use these additional signals to improve AoD estimation. Also, Algorithm~\ref{algo:mt} tracks the target when it moves in a 2D plane. 3D tracking can be enabled by extending Algorithm~\ref{algo:mt} to find the direction of paths in a 3D space.

The paper focuses on positioning alone since that is the hardest to provide. Orientation can be relatively easily obtained via IMUs already available on many devices~\cite{gearvr}.

In conclusion, we developed and implemented a WiFi-based motion tracking system and demonstrated its performance in different deployment scenarios. A commodity WiFi-based position tracking system would potentially enable VR on mobile devices and revolutionize number of applications that can be enabled on top of motion tracking. 

\section*{Acknowledgments}
\vspace{-2mm}
Manikanta Kotaru was supported by Thomas and Sarah Kailath Stanford Graduate Fellowship. The authors thank Prof. Gordon Wetzstein and Stanford Computational Imaging Group for valuable feedback on the paper and the anonymous reviewers for their insightful comments. 

\vspace{-2mm}
\section*{Glossary}
\begin{Glo}
\item \textbf{AoD (Angle of Departure): }The angle of a path with respect to a line joining two antennas at the source.
\item \textbf{AP: }A WiFi access point or router.
\item \noindent  \textbf{CSI (Channel State Information): }The received signal when the source transmits the reference signal.
\item \textbf{Frequency offset: }The difference in frequency of the reference signal used at two WiFi chips.
\item \textbf{Packet: }WiFi chunks data into units of efficient size for communication and each unit is called as a packet.
\end{Glo}

{\small
\bibliographystyle{ieee}
\bibliography{egbib}
}

\end{document}